\documentclass[twoside,11pt]{article}

\usepackage{mlhc}
\usepackage{times}
\usepackage{graphicx} 
\usepackage{subfigure} 
\usepackage{natbib}
\usepackage{floatrow}
\usepackage{amsmath}
\usepackage{float}
\usepackage{amsfonts}
\usepackage{graphicx}
\usepackage{wrapfig}
\usepackage{caption}
\usepackage{bm}

\usepackage{algorithm}
\usepackage{algorithmic}

\usepackage{hyperref}
\ShortHeadings{Improved Sepsis Detection with an MGP-RNN}{Futoma, Hariharan et al.}

\begin{document}

\title{An Improved Multi-Output Gaussian Process RNN with Real-Time Validation for Early Sepsis Detection}

\author{\name Joseph Futoma, Sanjay Hariharan, Katherine Heller \email jdf38,sh360,kh204@duke.edu \\
       \addr Department of Statistical Science\\
       Duke University, Durham, NC
       \AND
       \name Mark Sendak, Nathan Brajer \email mpd10,njb23@duke.edu \\
       \addr Institute for Health Innovation \\
       Duke University, Durham, NC
       \AND
       \name Meredith Clement, Armando Bedoya, Cara O'Brien \email me75,ab335,obrie028@duke.edu \\
       \addr Department of Medicine \\
		Duke University, Durham, NC
       \AND
       } 

\maketitle

\vspace{-.85in}

\begin{abstract}
Sepsis is a poorly understood and potentially life-threatening complication that can occur as a result of infection. Early detection and treatment improves patient outcomes, and as such it poses an important challenge in medicine. In this work, we develop a flexible classifier that leverages streaming lab results, vitals, and medications to predict sepsis before it occurs. We model patient clinical time series with multi-output Gaussian processes,  maintaining uncertainty about the physiological state of a patient while also imputing missing values. The mean function takes into account the effects of medications administered on the trajectories of the physiological variables. Latent function values from the Gaussian process are then fed into a deep recurrent neural network to classify patient encounters as septic or not, and the overall model is trained end-to-end using back-propagation. We train and validate our model on a large dataset of 18 months of heterogeneous inpatient stays from the Duke University Health System, and develop a new ``real-time'' validation scheme for simulating the performance of our model as it will actually be used. Our proposed method substantially outperforms clinical baselines, and improves on a previous related model for detecting sepsis. Our model's predictions will be displayed in a real-time analytics dashboard to be used by a sepsis rapid response team to help detect and improve treatment of sepsis.
\end{abstract}

\vspace{-.1in}

\section{Introduction}

Early detection of sepsis poses an important and challenging problem in medicine. Sepsis is a clinical complication from infections that has very high mortality and morbidity, and occurs when a person's immune system overreacts to the invasion of a microorganism and/or its toxin. The resulting inflammatory response can progress to septic shock, organ failure, and death unless it is intervened on early (\cite{sepsis}). However, even experienced providers can have significant difficulty identifying sepsis early and accurately, since the symptoms associated with sepsis can be caused by many other clinical conditions (\cite{Jones}). Actions such as early fluid resuscitation and administration of antibiotics within hours of sepsis recognition have been shown to improve outcomes (\cite{Ferrer}). Early intervention is crucial, as every hour that treatment is delayed after the onset of hypotension increases the risk of mortality from septic shock by 7.6\% (\cite{sepsis-death}).  Additionally, recent work found timely administration of a 3-hour care bundle was associated with lower in-hospital mortality across all septic patients (\cite{nejm-sepsis}), further emphasizing the need for fast and aggressive treatment.

With the widespread adoptions of electronic health records (EHRs), there  exists a wealth of data to inform predictions about when sepsis is likely to occur, which might help alleviate the lack of consistent early detection. Although some early warning scores that can use live data from the EHR to detect clinical deterioration exist, they are largely ad-hoc and not data-driven. One example is the National Early Warning Score (NEWS), which was developed to discriminate patients at risk of cardiac arrest, unplanned ICU admission, or death (\cite{NEWS}). Scores such as NEWS are typically broad in scope and were not designed to specifically target sepsis.  They are also very simple, as they use only a small number of variables (NEWS uses seven), and compare them to normal ranges to generate a single composite score. In assigning independent scores to each variable and using only the most recent value, they both ignore complex relationships between the variables and their evolution in time. It should not be surprising that implementation of such scores in clinical practice results in high alarm fatigue. An alarm based on NEWS was previously implemented in our university health system's EHR, but past work found that 63.4\% of alerts triggered were cancelled by the care nurse who received them. Our goal in this work is to develop a more flexible statistical model that uses all available information to make more accurate and timely predictions.

\begin{wrapfigure}{R}{0.64\textwidth}
\begin{center}
\centering
\includegraphics[width=\textwidth]{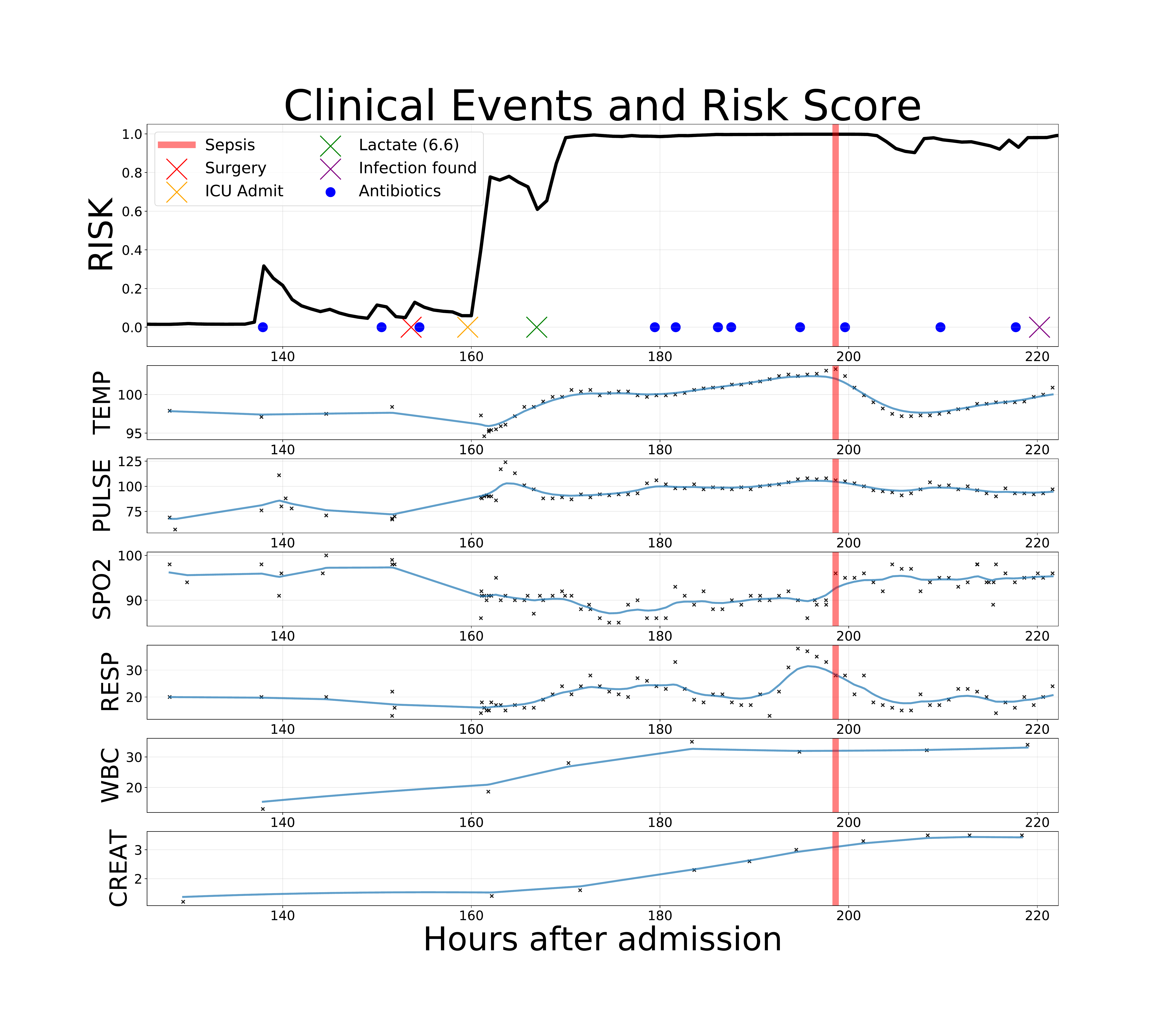}
\vspace{-0.2in}
\caption{This patient developed sepsis during a period of rapid deterioration in the ICU following an invasive cardiac surgery. However, our proposed model detected sepsis 17 hours before the first antibiotics were given and 36 hours before a definition for sepsis was met.}
\end{center}
\end{wrapfigure}

As a motivating example for our work, consider the patient data visualized in Figure 1, along with the risk scores generated by our proposed approach. This 37 year old female was initially admitted to the hospital for chest pains, and required an invasive cardiac surgery to clear a clot in her lungs. About six days passed between the time when she was admitted and when the surgery was to begin, during which she underwent many preoperative tests but was physiologically stable. However, following surgery she quickly destabilized and was admitted to the Intensive Care Unit (ICU). Shortly after her ICU admission, our model quickly predicted a high risk of sepsis due to her rapid deterioration, and after observing an abnormally high lactate (a common symptom of severe infection), the model became near certain that she was septic.  However, it was 17 hours after the model would have detected sepsis that her care team finally started treating her with antibiotics, and another 19 hours until a blood culture was drawn to ascertain the source of the infection. Fortunately, this patient fully recovered and was discharged a week later. Nonetheless, her care could have been better managed if her care team was aware of her sepsis earlier and prioritized treating it, which might have led to a faster recovery and shorter hospital stay.

The problem of identifying sepsis events from retrospective EHR data is  difficult. Unlike other clinical adverse events such as cardiac arrests or transfers to the ICU, the exact time at which sepsis ''starts" is not measurable. Suspected sepsis can be observed only indirectly, through abnormal labs and vitals, the administration of antibiotics, and the drawing of blood cultures to test for a suspected infection. This means the labels in our dataset for when sepsis occurred possess are noisy and not perfectly reliable. More generally, clinical time series presents additional modeling challenges, as typically measurements are obtained at irregular intervals and with frequent informative missingness, as measurements are often taken only if there is suspected problem. 

Our proposed model for detecting sepsis overcomes some of these limitations. The approach uses a Multiple-Output Gaussian Process (MGP) to de-noise and impute raw physiological time series data into a more uniform representation on an evenly spaced grid. The mean function of the MGP depends on medications, so the administration of different drugs affects the trajectory of the physiological time series. Latent function values from the process can then be fed into a deep recurrent neural network (RNN) classifier to predict how likely it is that a patient will acquire sepsis. The RNN also utilizes the informative missingness patterns from the clinical time series.  We train our model with data from a large cohort of heterogeneous inpatient encounters spanning 18 months extracted from our university health system EHR, and validate model performances with two methods, including a newly proposed ``real-time'' validation approach. 

\section{Related Works}

There are many previously published early warning scores for predicting clinical deterioration or other related outcomes. For instance, the NEWS score (\cite{NEWS}) and MEWS score (\cite{MEWS}) are two of the more common scores used to assess overall deterioration. The SIRS score for systemic inflammatory response syndrome was commonly used to screen for sepsis in the past (\cite{SIRS}), although it has been phased out by other scores designed for sepsis such as SOFA (\cite{SOFA}) and qSOFA (\cite{qSOFA}) in recent years. 

Within machine learning there has been much interest in modeling healthcare data. (\cite{trews}) present a simple Cox regression approach to prediction of sepsis using clinical time series data. The recent works  of (\cite{ForecastICU}) and (\cite{mihaela-nips}) are close in spirit to our application, as they develop models using clinical time series to predict more general deterioration as observed by admission to the ICU.  There are several related works that also utilize Gaussian processes in modeling multivariate physiological time series. For instance (\cite{MGP-phys1}) and (\cite{MGP-phys2}) use multitask Gaussian processes and (\cite{Chang2017}) use more complex multi-output Gaussian processes, with the focus in all on forecasting future vitals rather than predicting an event. Also relevant to our work is research using recurrent neural networks to classify clinical time series. In particular, (\cite{RNNtimeseries}) use Long-Short Term Memory (LSTM) RNNs to predict diagnosis codes given physiological time series from the ICU, and (\cite{RNN-HF}) use Gated Recurrent Unit RNNs to predict onset of heart failure using categorical time series of billing codes. In addition, (\cite{RNN_ts_miss}) and (\cite{Lipton_RNN_miss}) investigate patterns of informative missingness in physiological ICU time series with RNNs. Finally, our end-to-end technique to discriminatively learn both the MGP and classifier parameters builds off of our prior work (\cite{ICML_sepsis}), which in turn is based on (\cite{marlin2016}). 

\section{Proposed Method}

\begin{wrapfigure}{L}{0.5\textwidth}
\begin{center}
\centering
\includegraphics[width=\textwidth]{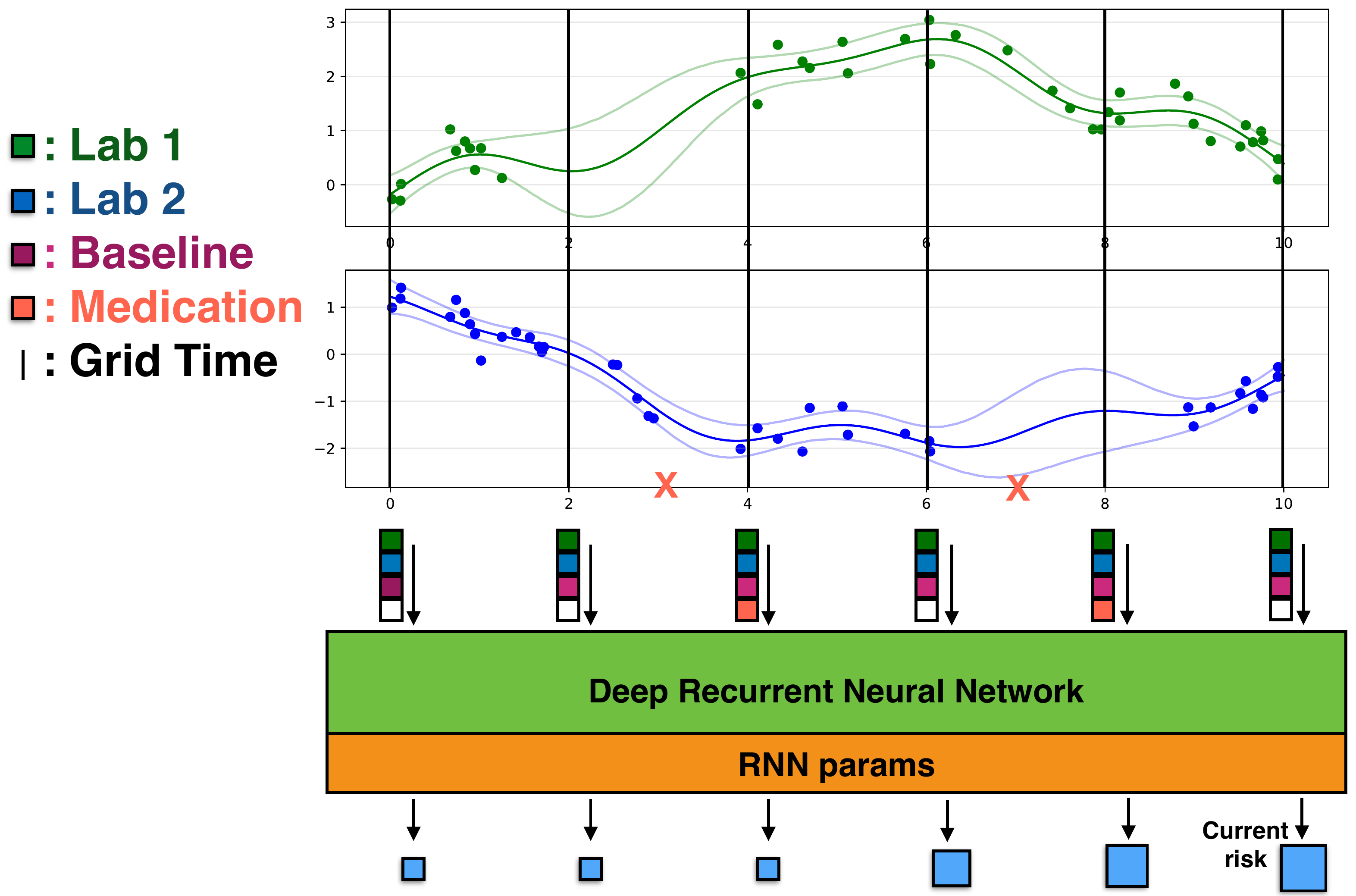}
\caption{Schematic for the overall method. At each grid time, latent function values for the clinical time series are fed into the RNN, along with baseline covariates and indicators for medications.} 
\end{center}
\vspace{-.2in}
\end{wrapfigure}

We view the problem of predicting sepsis as a multivariate time series classification. For a new patient admitted to the hospital, the goal is to provide updated risk scores for the probability that the patient will become septic, given the labs, vitals, medication, and baseline information available so far. Our proposed model improves upon our previous work in this area (\cite{ICML_sepsis}) that ties together a Multitask Gaussian Process with a Recurrent Neural Network classifier (MGP-RNN). We first introduce the original MGP-RNN framework, before highlighting several ways we have increased the flexibility of the MGP and improved the RNN classifier.

\subsection{The MGP-RNN Classifier}

The MGP-RNN Classifier is designed for irregularly spaced multivariate time series of variable length. The MGP is used to impute and de-noise raw clinical data into a more uniform representation on an evenly spaced grid so that it can be used by a downstream RNN classifier.  We now go into detail on each of these component pieces. Figure 2 is a schematic giving a high-level overview of the approach.


\subsubsection{Multi-output Gaussian Processes to De-noise and Impute}

Gaussian processes (GPs) are common models for irregularly spaced time series, as they naturally handle variable spacing and differing number of observations per series. An attractive property of GPs is that they maintain an estimate of uncertainty about the series at each point, which is important in this setting since clinical time series can be highly uncertain if there are few observations. Multi-output Gaussian Processes (MGPs) extend GPs to the setting of multivariate time series. Given $M$ outputs (physiological labs/vitals), the model is specified by a mean function $\{\mu_m(t)\}_{m=1}^M$ for each output, and a covariance function $K$. Letting $f_m(t)$ be a latent function representing the true value of output $m$ at time $t$, then $K(t,t',m,m') = \text{cov}(f_m(t),f_{m'}(t'))$, so $K$ specifies the covariances between different outputs across time. The actual observations are then distributed $y_m(t) \sim \mathcal{N}(f_m(t),\sigma_m^2)$ where $\{\sigma_m^2\}_{m=1}^M$ are noise variances.  A common simplifying assumption introduced in (\cite{MGP}) and later by (\cite{ICML_sepsis}), (\cite{MGP-phys1}), and many others is that this kernel is separable, i.e. $K(t,t',m,m') = K^M_{mm'}k^t(t,t')$, so the covariances between inputs and outputs are modeled separately. $k^t$ is a shared correlation function across time and $K^M$ is a $M \times M$  positive-definite matrix specifying the covariances between outputs. In practice we use the Ornstein-Uhlenbeck covariance function, $k^t(t,t') = e^{-|t-t'|/l}$.

Although we relax the separable kernel assumption later, it has the convenient property that for a complete time series, the $MT \times MT$ (assuming $T$ time points) covariance matrix for observations $(y_{11}, \dots, y_{1T}, y_{21}, \dots, y_{2T}, \dots, y_{MT})$ is  expressed as $\Sigma = K^M \otimes K^T + D \otimes I$, where $y_{mj}$ is the observed value for variable $m$ at the $j$'th time $t_j$, $\otimes$ is the Kronecker product, $K^T$ is a $T \times T$ correlation matrix between observation times as specified by $k^t$, and $D$ is a diagonal matrix of the noise variances $\{\sigma_m^2\}_{m=1}^M$. In practice, only a subset of the $M$ series are observed at each time, so $\Sigma$ only needs to be computed at the observed variables. Another common assumption is that the MGP has zero mean (i.e. each output has been centered), although we will also relax this later.

We use the MGP to handle the irregular spacing and missing values in the raw clinical data and output a more uniform representation to feed into a downstream classifier. To accomplish this, let $\bm{x}$ be a vector of evenly spaced points in time (in practice, $x_1=0$ is admission time, with future times spaced an hour apart) that will be shared across all encounters. The MGP provides a posterior distribution for the $X \times M$ matrix $\bm{Z}$ (where $X = |\bm{x}|$) of the latent true $M$ time series values at $X$ evenly spaced grid times for a particular encounter. This conditional normal posterior importantly maintains uncertainty about the true function values, while also de-noises and imputes each variable on a grid that makes it possible to use as input to a black box classifier.

\subsubsection{Long Short Term Memory RNNs to Classify}

Following (\cite{ICML_sepsis}) we learn a classifier that takes the latent function values $\bm{z} \equiv \text{vec}(\bm{Z})$ at shared reference grid times $\bm{x}$ as inputs, where $\bm{z} \sim N(\mu_{z}, \Sigma_{z}; \bm{\theta})$ is distributed according to the MGP posterior, which depends on hyperparameters $\theta$. We use deep recurrent neural networks as our classifier, a natural choice for learning flexible functions that map variable-length input sequences to a single output, in particular using the Long Short Term Memory (LSTM) architecture (\cite{LSTM}). For encounter $i$, at each time $x_{ij}$ inputs $\bm{d}_{ij}$ will be fed into the network, consisting of: $M$ latent function values $\bm{z}_{ij}$, $B$ baseline covariates $\bm{b}_i$, and $P$ counts $\bm{m}_{ij}$ of medications administered between $x_{ij}$ and $x_{i,j-1}$, i.e. $\bm{d}_{ij} = [\bm{z}_{ij}^\top,\bm{b}_i^\top,\bm{m}_{ij}^\top]^\top$. The RNN learns complex time-varying interactions among baseline covariates,  labs and vitals, and medications. 

However, $\bm{z}_i$ are latent variables and not observed directly. Thus the RNN classification output $f(\bm{D}_i;\bm{w})$ (mapping a matrix of inputs $\bm{D}_i$ to a probability, parameterized by $\bm{w}$), and hence the model loss function $l(f(\bm{D}_i; \bm{w}), o_i)$, is stochastic (where $o_i$ is the true label).  Given $\bm{z}_i$, learning the classifier would involve finding parameters $\bm{w}$ to minimize this loss; since $\bm{z}_i$ is unobserved, we instead optimize the expected loss $\mathbb{E}_{\bm{z}_i \sim N(\mu_{z_i}, \Sigma_{z_i}; \theta)}[l(f(\bm{D}_i; \bm{w}), o_i)]$ with respect to the MGP posterior for $\bm{z}_i$. The overall learning problem is then to minimize the expected loss over the full dataset of $N$ encounters: $\bm{w}^*, \bm{\theta}^* = \text{argmin}_{\bm{w},\bm{\theta}} \sum_{i=1}^N \mathbb{E}_{\bm{z}_i \sim N(\mu_{z_i}, \Sigma_{z_i}; \theta)}[l(f(\bm{D}_i; \bm{w}), o_i)]$.

We optimize the loss with stochastic gradient descent using ADAM  (\cite{ADAM}).  Since the expected loss is intractable, we approximate it with Monte Carlo samples by taking draws of $\bm{z}_i$ from its MGP posterior.  We compute gradients of the loss with respect to the RNN parameters $\bm{w}$ and the MGP parameters $\bm{\theta}$ with the reparameterization trick, as $\bm{z}_i= \bm{\mu_{z_i}} + \bm{R}_i \bm{\xi}_i$, where $\bm{\xi}_i \sim N(0,I)$ and $\bm{R}_i$ is a matrix such that $\bm{\Sigma_{z_i} = R_iR_i}^\top$ (\cite{reparam}).  We use the Lanczos method (\cite{marlin2016}) to speed computation associated with drawing samples of $z_i$, as this involves drawing from a potentially very large multivariate normal.  Since every step in these algorithms are differentiable we can  use backpropagation to compute gradients. 

This is the methodology used in (\cite{ICML_sepsis}): a zero-mean, separable MGP to denoise and impute, fed into an LSTM to classify. We now highlight improvements to this base method.

\subsection{Increasing Flexibility of the Multitask Gaussian Process}

We increase the flexibility of the MGP in two ways. First, we incorporate medication effects so that the mean of each physiological variable depends on the administration of past drugs. Second, we relax the assumption that the kernel function be separable.

\subsubsection{Incorporating Medication Effects}

We relax the zero mean function assumption, and let the mean depend on previous administration of medications. We let the prior mean function $\mu_m(t)$ for lab/vital $m$ at time $t$ be expressed as $\mu_m(t) = \sum_{p=1}^P \sum_{t_p < t} f_{pm}(t-t_p)$, where $f_{pm}$ is a function that specifies the effect medication $p$ has on lab/vital $m$, and $\{t_p\}$ is the times drug $p$ was given.  We use $f_{pm}(t) = \sum_{l=1}^L \alpha_{lpm} e^{-\beta_{lpm} t}$ ($\beta_{lpm}>0$), a flexible family of curves that allows for effects to occur on different length-scales.  Each time a new drug is given, the mean function spikes according to $f_{pm}$. We set $L=3$.

\subsubsection{Sum of Separable Kernel Functions}

We relax the separable covariance function by considering a sum of $Q$ separable covariance functions, each with their own parameters, $K_q^M$ and  $l_q$. The resulting covariance matrix can be written as $\Sigma = \sum_{q=1}^{Q} K_q^M \otimes K_q^T + D \otimes I$.  This is a more flexible family of covariance functions, and no longer forces all output variables to share the same temporal correlation structure (\cite{MGP_survey}, \cite{collab_MGP}). This model is also equivalent to the well-known Linear Model of Coregionalization (\cite{Geostats}). We found $Q=3$ worked well in practice.

\subsection{Improving the RNN Classifier}

We improve the RNN classifier in two ways. First, we use target replication to increase the signal at the end of the series and make learning easier. Second, we use the pattern of missingness in the raw labs/vitals to improve predictions.

\subsubsection{Target Replication}

Instead of the loss function depending only on the output at the final time step, following (\cite{RNNtimeseries}) we use target replication so the loss function depends on the outputs of the RNN at multiple time points.  This helps to alleviate issues with our imprecise labels for the true time of sepsis, as we can simply label multiple time points near a given time of sepsis. In practice, we use target replication by labelling additional times from 2 hours prior to 6 hours after a sepsis event.

\subsubsection{Utilizing Missingness Patterns}

We increase the flexibility of our approach by directly modeling the patterns of missing data in the physiological variables, similar to the ideas in (\cite{Lipton_RNN_miss}).  To each input vector into the RNN, containing latent physiological function values from the MGP, baseline covariates, and medications administered, we append a binary vector denoting which labs have been sampled since the last grid time.  This will allow the RNN to model complicated interactions between the missingness patterns in the time series variables, along with the learned values of the variables themselves, and the baseline covariates and meds. This additional information can be very useful, as many labs are only ordered when there is a suspected problem.

\section{Experiments}

\subsection{Data Description}

Our training dataset consists of 51,697 inpatient admissions from our university health system spanning 18 months, extracted directly from our Epic EHR. There are $M=34$ continuous-valued physiological variables, of which 6 are vitals (e.g blood pressure), and 28 are laboratory values (e.g. lactate). There are $B=35$ covariates collected at baseline, of which 29 are comorbidities (e.g. history of cardiac disease), in addition to race, gender, age, and whether the admission was a transfer, was urgent, or was an emergency. Finally, we have the times of administration of $P=8$ medication classes (e.g. antibiotics). The patient encounters range from very short admissions of only a few hours to extended stays lasting many weeks, with the mean length of stay at 121.7 hours, with a standard deviation of 108.1 hours. The resulting population is very heterogeneous as there was no specific inclusion or exclusion criteria. This makes the cohort representative of the clinical setting in which our method will be used, as the goal is to apply it broadly throughout the hospital.

For encounters that resulted in sepsis, we used a well-defined clinical definition to assess the first time at which sepsis is suspected to have been present. The criteria was: at least two persistently abnormal vitals signs (SIRS score of at least 2/4), a blood culture drawn for suspected infection, and at least one abnormal lab indicating early signs of organ failure\footnote{Our criteria most closely matches the ``Sepsis-2'' definition for severe sepsis (\cite{Sep2}).  Although a ``Sepsis-3'' definition was recently released (\cite{qSOFA}), it tends to identify sicker patients with higher mortality, compared with ``Sepsis-2'', and its adoption is not yet standard.  In order to identify more patients potentially at risk of sepsis, we used the older definition.  However, our 	methodology is general and could easily be applied to a similar dataset with a different definition for sepsis.}.  The overall rate of sepsis in the dataset was 21.4\%, with each encounter associated with a binary label of whether the patient acquired sepsis, along with a time our sepsis definition was met.

\subsection{Experimental Setup}

We trained our method on 80\% of the dataset, setting aside 10\% for validation to select hyperparameters and the remaining 10\% for final evaluation. For all RNNs we used a 2 layer LSTM with 64 hidden units per layer. We used $L_2$ regularization on the weights  and early stopping to guard against overfitting. We train all models using stochastic gradient descent with minibatches of 100 encounters and learning rate of 0.001. Our methods are implemented in Tensorflow\footnote{\url{https://github.com/jfutoma/MGP-RNN}}.

\subsubsection{Case Control Matching}

For septic patients we retain data up until 6 hours after sepsis was acquired (for target replication). For non-septic patients, it is not very clinically relevant to include all data up until discharge, and compare predictions about septic encounters shortly before sepsis with predictions about non-septic encounters shortly before discharge. This task would be too easy, as the controls before discharge are likely to be  clinically stable. To make the learning problem more challenging and improve the generalizability of the model, we use a form of case-control matching. The model is then trained to label sepsis encounters around the time of sepsis, and to label control encounters at some time mid-encounter. In particular, we first match each sepsis encounter to 4 non-sepsis encounters (this roughly maintains the actual sepsis rate of around 20\%) with similar lengths of stay and baseline covariates. Then, we mark a ``prediction time" for each control encounter to be at the same fraction of its length of stay as sepsis was during its matched sepsis encounter (e.g. if sepsis occurred at 25\% through an encounter, for each matched control we use the time 25\% through). To train and evaluate our models, we now use data until the time of sepsis plus six additional hours for sepsis cases, and this ''prediction time" plus six additional hours for the controls. This is a more realistic problem, since the non-sepsis encounters may not be near discharge now and will be less clinically stable, and the model will must learn what differentiates them from sepsis cases.

\subsubsection{Methods Compared}

We compare a number of variants of our method against several simpler models and clinical baselines. Our base method, denoted ``Base MGP-RNN", is the model from (\cite{ICML_sepsis}) with none of the extensions from Sections 3.2 and 3.3.  The method denoted ``Target Replication" adds target replication to this from Section 3.3.1, using labels from 2 hours prior to sepsis until 6 hours after sepsis. The method ``SoS kernel" adds to this by using a sum of $Q=3$ separable kernels as described in Section 3.2.2. ``Medication effect" further adds to this by learning a treatment-response curve for the mean function of the MGP in each dimension, as in Section 3.2.1. Finally, ``Missingness Indicators" uses all the previous extensions and also feeds indicator vectors for when each physiological variable is measured into the RNN, from Section 3.3.2.

\begin{figure}[ht]
\centering
\includegraphics[width=16cm]{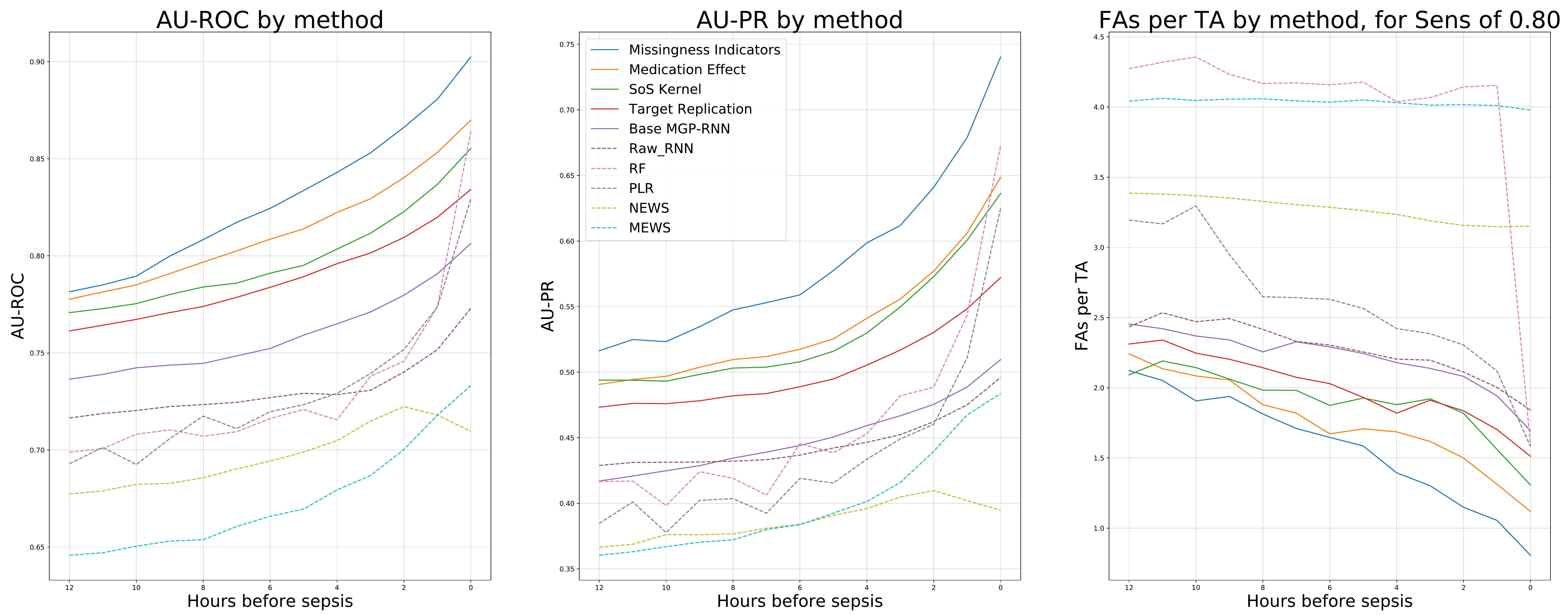}
\caption{Results from Matched Lookback validation scheme.  Left: AU-ROC as a function of number of hours before sepsis or the matched ``prediction time".  Middle: AU-PR as a function of hours before sepsis. Right: For a fixed sensitivity of 80\%, the number of false alarms per true alarm as a function of hours before sepsis.}
\end{figure}

Our strongest baseline method, ``Raw RNN", consists of the same network architecture as the MGP-RNN, but instead uses the mean value of each lab/vital in hourly windows, and for periods with missing data the most recent value is carried forward (we use the median for all values of that lab/vital across all encounters if there was no value to carry forward). We also compare with a Lasso logistic regression (``PLR") and random forest (``RF") fit to the same data as the Raw RNN and with the same imputation strategy. Finally, the NEWS and MEWS scores were used as clinical baselines.

\subsection{Evaluation Schemes and Results}

We use the area under the ROC curve and the area under the Precision Recall curve as evaluation metrics to compare how each method's performance differs.  We also examine the number of false alarms per true alarm for each method, a metric directly related to precision. We first introduce what we call a ``Matched Lookback Validation" scheme and present results from it, and then introduce a ``Real-Time Validation" scheme and present additional results.

\subsubsection{Matched Lookback Validation}

\begin{wrapfigure}{R}{0.4\textwidth}
\centering
\includegraphics[width=1.1\textwidth]{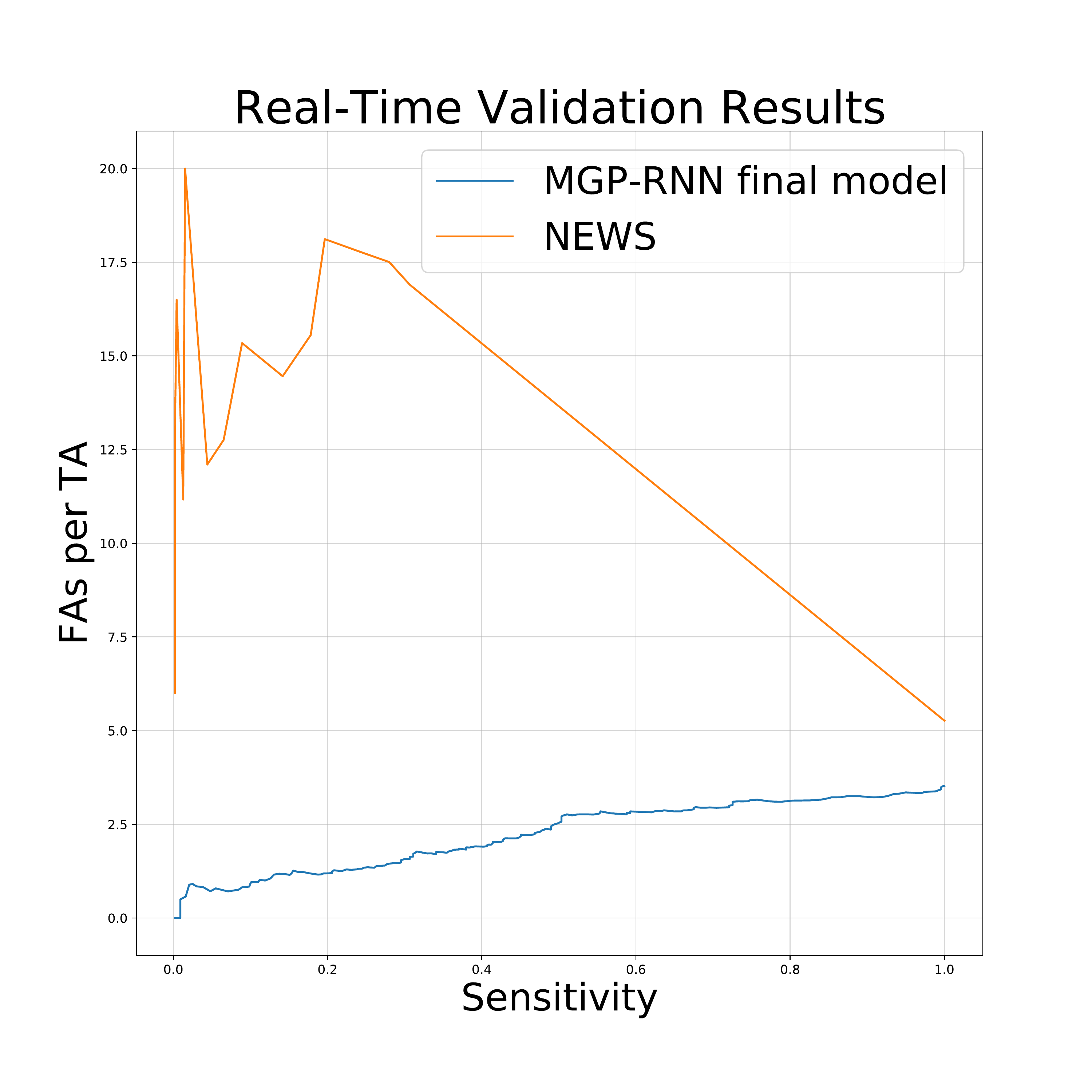}
\caption{False alarms per true alarm as a function of sensitivity using the real-time validation scheme, for our final model and for NEWS. Our model demonstrates drastic reductions in false alarms across all sensitivities.}
\end{wrapfigure}

In this validation strategy, we align matched encounters at either time of sepsis or the ``prediction time" for controls, and see how model performance degrades as we make predictions a fixed number of hours in advance of this time.  That is, we compare how well the methods discriminate between sepsis and control using all data up through the actual time of sepsis / ``prediction time", then up until 1 hour before, and so on, up until 12 hours in advance.  This will give a sense for how far in advance we can reliably predict sepsis.

Figure 3 shows the results from this validation mechanism. It is clear that the various MGP-RNN methods substantially outperform both the clinical baselines and the other baseline models. The extensions presented to the Base MGP-RNN all improve its performance by a modest margin. The most complete model with all the extensions considered consistently outperformed all other methods for all of the metrics we considered. The number of false alarms per true alarm (right pane of Figure 3) is the most clinically useful metric.  At 4 hours prior to sepsis, our best model only had about 1.4 false alarms per true alarm, at a very high sensitivity of 80\%; compare this to the 2.2 false alarms the base MGP-RNN has, and the 3.2 false alarms that the NEWS score that was previously implemented at our hospital had.  

\subsubsection{Real-Time Validation}

A criticism of the previous validation mechanism is that it requires alignment of patients by when their sepsis or ``prediction time" is, and this will not actually be known in practice when actually used.  To alleviate this, we also validate our approach in a more ``real-time" setting.  For each encounter, we first generate a ``real-time" risk score at each hour in time, i.e. using only data up until that point. Then, we choose a risk threshold, so that a risk above that fires an ``alarm". We then construct a confusion matrix across all encounters for this threshold, using the following logic.  A false negative occurs either when an encounter resulted in sepsis but an alarm was never fired, or the alarm would have fired after sepsis already occurred.  A true negative occurs if an alarm never fires for a control encounter. If an alarm fires at any point for a control encounter, a false positive results.  Finally, if an alarm fires for a sepsis encounter between 0 and 48 hours before sepsis, we count it as a true positive.  But, if it fires more than 48 hours in advance, it is a false positive (we don't want to count situations where an alarm fires and sepsis happened many days or weeks later).  Following this procedure, we can construct confusion matrices for a variety of risk thresholds and use them to produce metrics such as ROC and Precision-Recall curves that typically do not depend on time. 

For simplicity, we only used real-time validation to compare our best model (``Missing indicators", from the lookback results) to the NEWS score, since the NEWS score was previously used in practice, while our model is currently being used. Figure 4 shows the number of false alarms per true alarm across all sensitivities for the two methods. Clearly, our proposed method offers large reductions in the number of false alarms across all sensitivities, and should substantially decrease the high alarm fatigue associated with NEWS.

\section{Conclusions and Clinical Significance}

\begin{wrapfigure}{R}{0.4\textwidth}
\centering
\includegraphics[width=\textwidth]{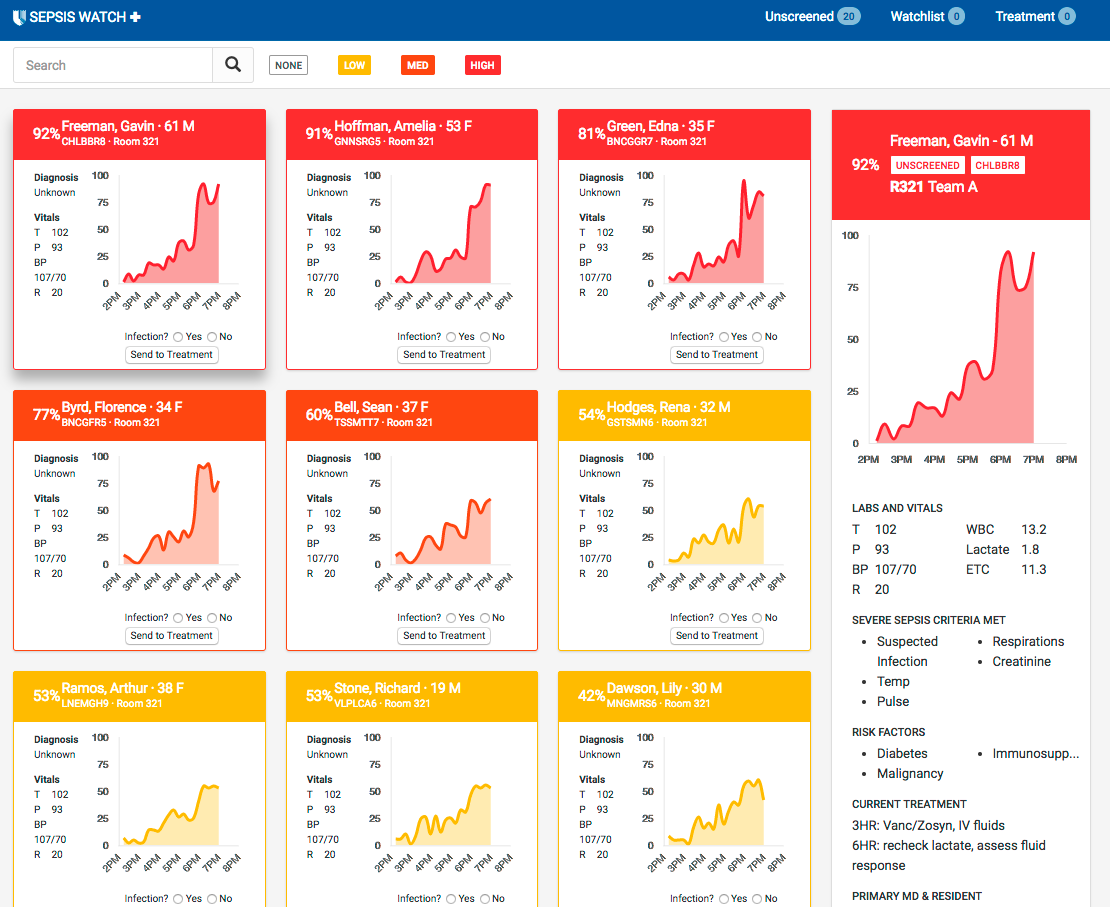}
\caption{Screenshot of analytics dashboard (with fake data) that will be used to visualize our model's predictions, to be used by  a sepsis rapid response team.}
\end{wrapfigure}

In this work we presented an improved methodology for early detection of sepsis, building on a related previous work. We find that our proposed methods outperform strong baselines and several clinical benchmarks, and offer substantial improvements over the model they build off of.  However, there are several many avenues for future work to improve the model. One obvious direction is to improve interpretability of the model so that it is possible to see which inputs at which times contributed the most to the risk score. We are actively investigating this by adding an attention mechanism to the RNN, e.g. along the lines of (\cite{RETAIN}). Furthermore, better methods to capture heterogeneity in this population by clustering or learning latent subpopulations with similar clinical statuses might help to improve overall performance.  Exploring other types of flexible black box classifiers such as recurrent variational auto-encoders (e.g. \cite{RNN-VAE}) may also improve the model's performance by better accounting for uncertainty in the classifier parameters.  Finally, it would also be interesting to combining this work with a reinforcement learning approach to learn not only how to detect sepsis early but also optimal treatment strategies. 

Due to the importance of this problem in medicine, our work has the potential to have a high impact in actual clinical practice. In Figure 5 we present a snapshot of an analytics dashboard that is currently being deployed at our hospital system's wards. The tool will be used to display the predictions of our model to predict sepsis to clinicians and nurses on a rapid response team specifically designed to facilitate early detection of sepsis.  The application and our model's risk scores will help ensure that early interventions for treatment of sepsis can be started faster for the highest risk patients. Use of our model compared to the NEWS score previously used should dramatically reduce alarm fatigue, and will hopefully both improve patient outcomes and reduce overall burden on the providers. Although in this paper our emphasis was on early detection of sepsis, the methods could be used with little modification to detect other clinical adverse events, such as cardiac arrest or admission to the ICU. 

\bibliography{CR.bib}

\end{document}